%% file: acl_latex.tex
\pdfoutput=1

\documentclass[11pt]{article}

\usepackage[final]{acl}

\usepackage{times}
\usepackage{latexsym}

\usepackage[T1]{fontenc}

\usepackage[utf8]{inputenc}

\usepackage{microtype}

\usepackage{inconsolata}

\usepackage{inconsolata}
\usepackage{amsmath}
\usepackage{amssymb}
\usepackage{mathtools}
\usepackage{amsthm}
\usepackage{tabularx}
\usepackage{enumitem}
\usepackage{multirow}
\usepackage{multicol}
\usepackage{makecell}
\usepackage{booktabs}
\usepackage{subfigure}

\title{Adaptive Feature-based Low-Rank Compression of Large Language Models via Bayesian Optimization}

\author{Yixin Ji$^{1}$\thanks{\; Equal Contribution.}, Yang Xiang$^1$\footnotemark[1], Juntao Li$^{1}$\thanks{\; Corresponding author.}, \\{\bf Qingrong Xia$^2$}, {\bf Zi Ye$^2$}, {\bf Xinyu Duan$^2$}, {\bf Zhefeng Wang$^2$}, {\bf Kehai Chen$^3$}, {\bf Min Zhang$^{1,3}$} \\  
 $^{1}$School of Computer Science and Technology, Soochow University \\
 $^{2}$Huawei Cloud, China \\
 $^{3}$Harbin Institute of Technology, Shenzhen \\
 \texttt{\{jiyixin169,baldwin021129\}@gmail.com}; \\
  \texttt{\{ljt,minzhang\}@suda.edu.cn} \\
 }

\begin{document}
\maketitle
\begin{abstract}
In recent years, large language models (LLMs) have driven advances in natural language processing. Still, their growing scale has increased the computational burden, necessitating a balance between efficiency and performance. Low-rank compression, a promising technique, reduces non-essential parameters by decomposing weight matrices into products of two low-rank matrices. Yet, its application in LLMs has not been extensively studied. The key to low-rank compression lies in low-rank factorization and low-rank dimensions allocation. To address the challenges of low-rank compression in LLMs, we conduct empirical research on the low-rank characteristics of large models. We propose a low-rank compression method suitable for LLMs. This approach involves precise estimation of feature distributions through pooled covariance matrices and a Bayesian optimization strategy for allocating low-rank dimensions. Experiments on the LLaMA-2 models demonstrate that our method outperforms existing strong structured pruning and low-rank compression techniques in maintaining model performance at the same compression ratio.\footnote{The implementation code and model checkpoints are available at \url{https://github.com/Dereck0602/Bolaco}.}
\end{abstract}

\input{article/1-introduction}

\input{article/2-preliminary}

\input{article/3-method}

\input{article/4-experiments}

\input{article/5-results}

\input{article/6-related_work}

\input{article/7-conclusion}

\section*{Limitations}
Although our proposed Bolaco has made significant progress in low-rank compression for LLMs, there are still some limitations:
\begin{itemize}[leftmargin=*]
    \setlength{\itemsep}{0pt}
    \setlength{\parskip}{0pt}
    \item Due to computational resource constraints, we only conduct thorough experiments on two commonly used LLaMA 2 models, lacking investigation into larger models (such as LLaMA 2-70B), other architectures (such as the OPT and T5 families), and multimodal models.
    \item To improve the efficiency of Bayesian optimization, we reduced the parameter dimensions by sharing parameters of different types and layers using low-rank dimensions. This may limit the potential performance of the model. We plan to use more advanced methods to find better low-rank allocation while maintaining flexibility.
    \item Compared to state-of-the-art structured pruning, low-rank compression falls short in highly compressed language models, but exhibits better zero-shot performance on downstream tasks. These observations inspire us to investigate how to effectively combine these two approaches to capitalize on their advantages in the future.
\end{itemize}

\section*{Acknowledgments}
We want to thank all the anonymous reviewers for their valuable comments. This work was supported by the National Science Foundation of China (NSFC No. 62206194, 62276077, and U23B2055), the Natural Science Foundation of Jiangsu Province, China (Grant No. BK20220488), Young Elite Scientists Sponsorship Program by CAST (2023QNRC001), and Huawei Cloud.

\bibliography{acl_latex}

\clearpage
\appendix

\section{More Experiments on Low-rank Sensitivity}
\label{sec:sensitive}
\begin{figure*}[t]
\centering
\subfigure[Low rank sensitivity of individual layer on LLaMA-v2-7b.]{
\begin{minipage}[t]{\linewidth}
\includegraphics[scale=0.55]{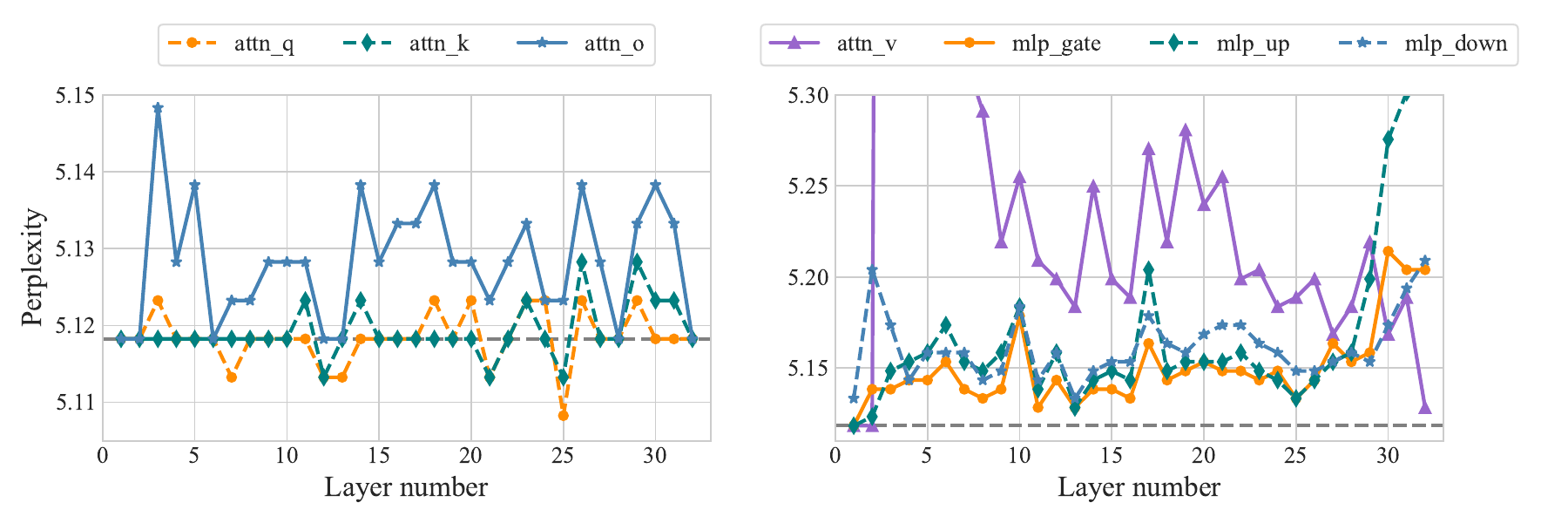}
\end{minipage}
}\\
\subfigure[Sensitivity of different types of layers to low-rank compression on the LLaMA-v2-7b-chat.]{
\begin{minipage}[t]{\linewidth}
\includegraphics[scale=0.55]{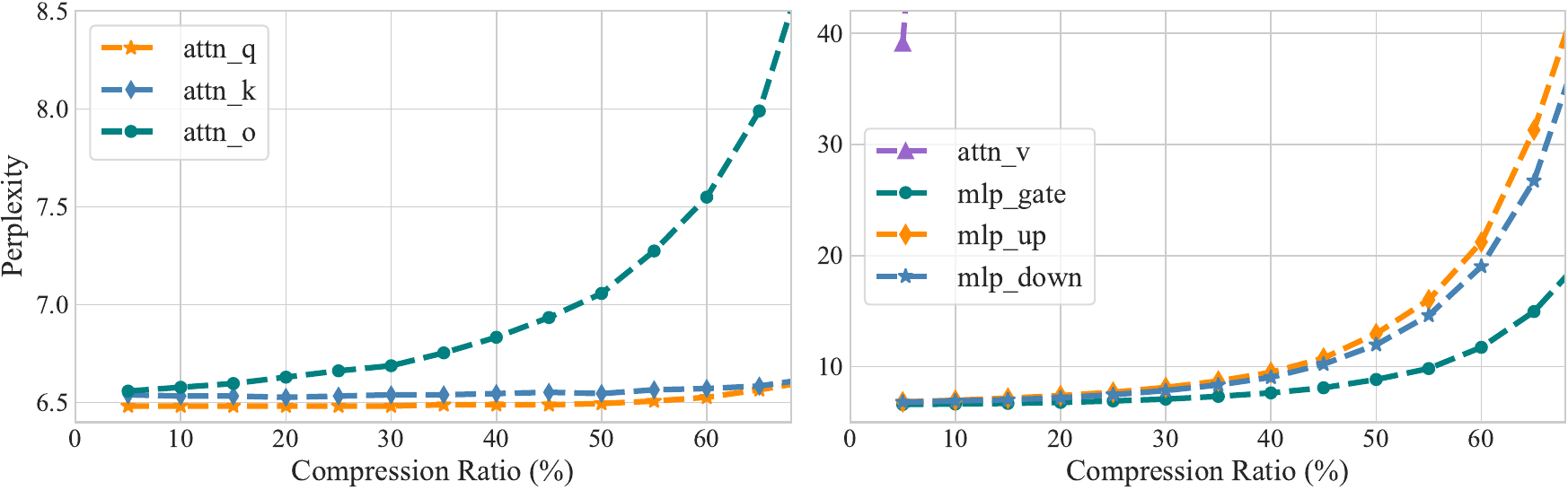}
\end{minipage}
}\\
\subfigure[Sensitivity of different types of layers to low-rank compression on the OPT-6.7b.]{
\begin{minipage}[t]{\linewidth}
\includegraphics[scale=0.55]{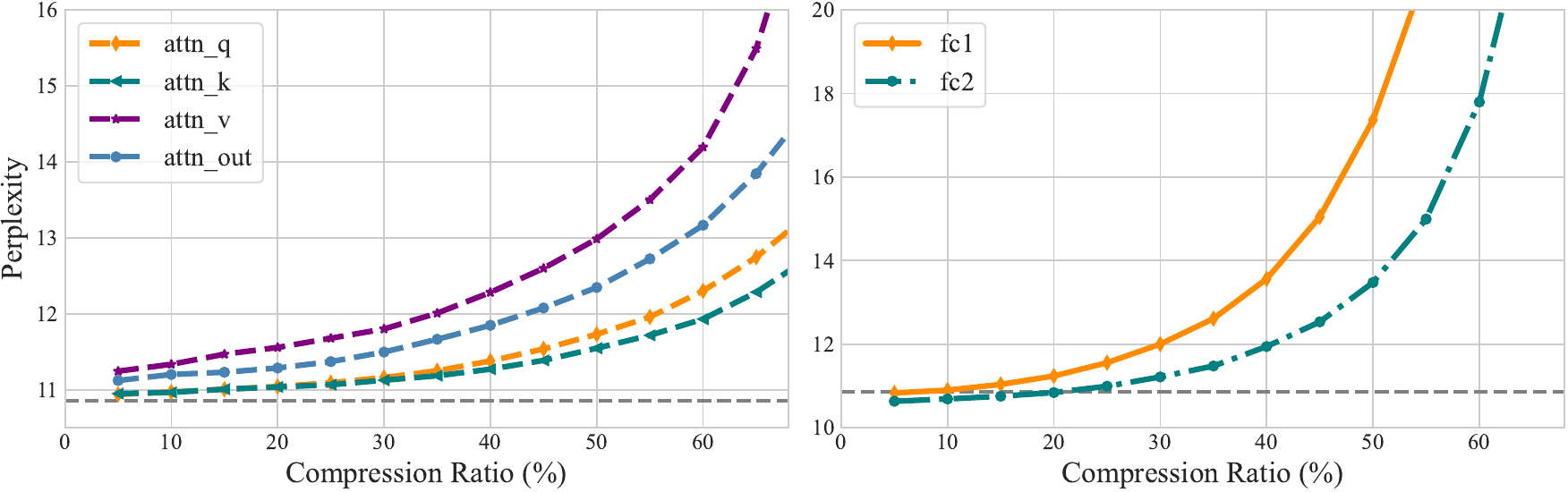}
\end{minipage}
}\\
\captionsetup{font={small}}
\caption{More results on low-rank sensitivity.}
\label{fig:moresen}
\end{figure*}
As shown in Figure \ref{fig:moresen}. We further reduce the parameters by 50\% on LLaMA-v2-7b by low-rank compression for each layer and test its perplexity on WikiText2.
We observe that the low-rank sensitivity varies significantly across different types of parameters. Compression of $attn\_q$ and $attn\_k$ seemingly has negligible impact on overall performance across all layers. In contrast, the upper layers of $mlp$ are more sensitive compared to the lower and middle layers.
We also conduct experiments on LLaMA-7b-chat and OPT-6.7b and find significant variability between all the different types of parameters. However, for OPT-6.7b, the differences between them are less pronounced than for LLama, especially for $attn\_v$, which does not show an explosive increase in perplexity.

\section{Baselines}
\label{sec:baseline}
\noindent{\textbf{LLM-Pruner} }\cite{ma2023llmpruner} is a dependency-aware one-shot structured pruning method.
It evaluates the importance of each structure through a first-order Taylor expansion and prunes the structures with the lowest scores. 
After pruning, it uses LoRA post-training to recover performance.

\noindent{\textbf{FLAP}} \cite{an2023fluctuationbased} is an one-shot retraining-free structured pruning method.
It utilizes a fluctuation-based metric to measure the impact of pruning on features and employs a bias term to compensate for the pruning loss.

\noindent{\textbf{SliceGPT}} \cite{ashkboos2024slicegpt} is a post-training sparsification method.
It replaces each weight matrix with a smaller matrix, reducing the embedding dimension of the network.

\noindent{\textbf{LoRD}} \cite{kaushal2023lord} is a naive feature-based low-rank compression method for code LLMs.
It does not take into account the low-rank allocation of varying parameters.
We migrate it to generic LLaMA-family LLMs.

\noindent{\textbf{ASVD}} \cite{yuan2023asvd} is a training-free SVD-based LLM compression method.
It manages activation outliers by scaling the weight matrix based on the activation distribution.

\section{Implementation Details}
\label{sec:details}
For LoRD, due to the absence of reference settings for its application on the LLaMA, we manually search a good low-rank allocation for it.
At the 20\% compression ratio, we do not compress $attn\_v$, and reduce the parameter count of $attn\_q/k$ by 30\%, with a 20\% reduction in the remaining parameters.
At the 30\% ratio, we reduce the parameter count of $attn\_q/k$ by 45\%, with a 30\% reduction in the remaining parameters except $attn\_v$.

At the post-training stage, we only add fine-tunable low-rank matrices for the compressed parameters.
We set the low-rank dimension $r^\prime=256$, the learning rate is 2e-3, and the batch size is 64.

\input{table/add_result}

\input{table/mistral}

\section{Discussion on compute intensive about Bolaco}
The computational cost of our method is divided into three parts:

\noindent{\textbf{PCA decomposition}}
Our method requires only one PCA decomposition of the obtained representations and truncates them according to the assigned rank during the rank allocation process to generate various compressed models. The computational cost here is the same as that of the existing low-rank decomposition method LORD. 

\noindent{\textbf{Obtaining evaluation results}}
In our experiments, the validation set we selected is not large, about 34k tokens, so the validation process takes less time, and the total validation time spent by llama-2-7b is about 40-45min on a 40G A100.

\noindent{\textbf{Bayesian Optimization}}
The computational time for 50 epochs of Bayesian optimization is approximately 45-50 minutes, which is considered acceptable in practical applications. Compared to iterative pruning, Bayesian optimization is more memory-efficient, as it only requires the memory overhead of forward propagation without storing gradients, momentum, or other optimizer states. Furthermore, as discovered in Section 5.5, the low-rank configurations optimized on a base model can be transferred directly, or with few rounds of Bayesian optimization, to variant models with the same architecture. It implies that we can quickly obtain a well-performing, low-rank compressed model for fine-tuned LLMs on different datasets in practice.

\section{Statistics of the Compressed
Model}

\begin{figure*}[htbp]
	\centering
	\subfigure[The perplexity of WikiText2 on LLaMA-v2-7b]{
		\begin{minipage}[b]{0.4\textwidth}
			\includegraphics[width=1\textwidth]{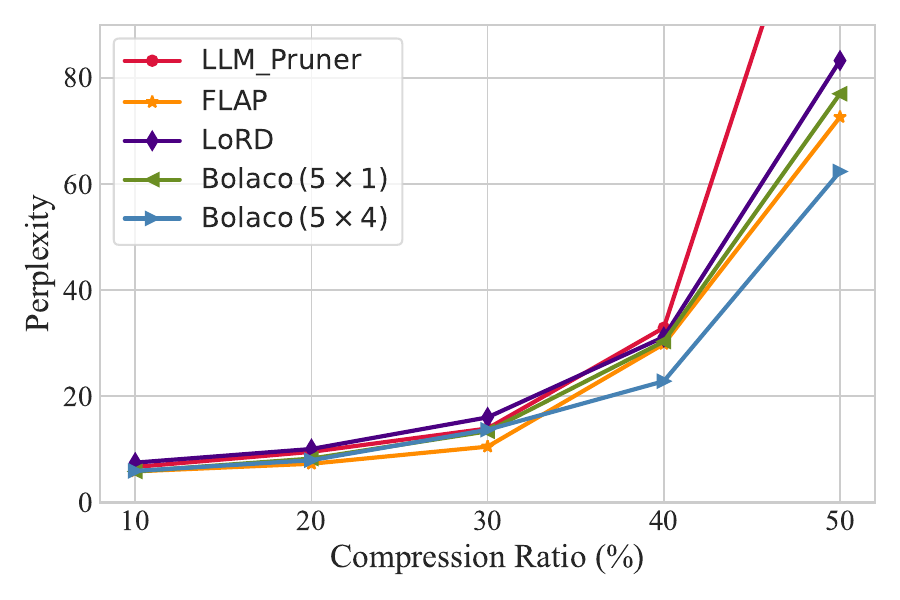} 
		\end{minipage}
		\label{fig:grid_4figs_1cap_4subcap_1}
	}
        \subfigure[The perplexity of WikiText2 on LLaMA-v2-13b]{
    		\begin{minipage}[b]{0.4\textwidth}
   		 	\includegraphics[width=1\textwidth]{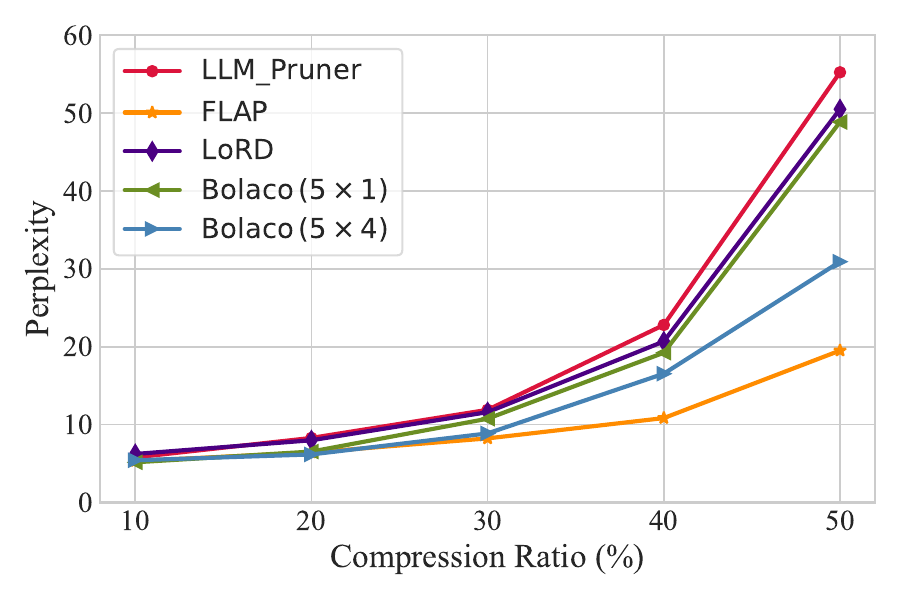}
    		\end{minipage}
		\label{fig:grid_4figs_1cap_4subcap_2}
    	}
	\\
	\subfigure[The perplexity of C4 on LLaMA-v2-7b]{
		\begin{minipage}[b]{0.4\textwidth}
			\includegraphics[width=1\textwidth]{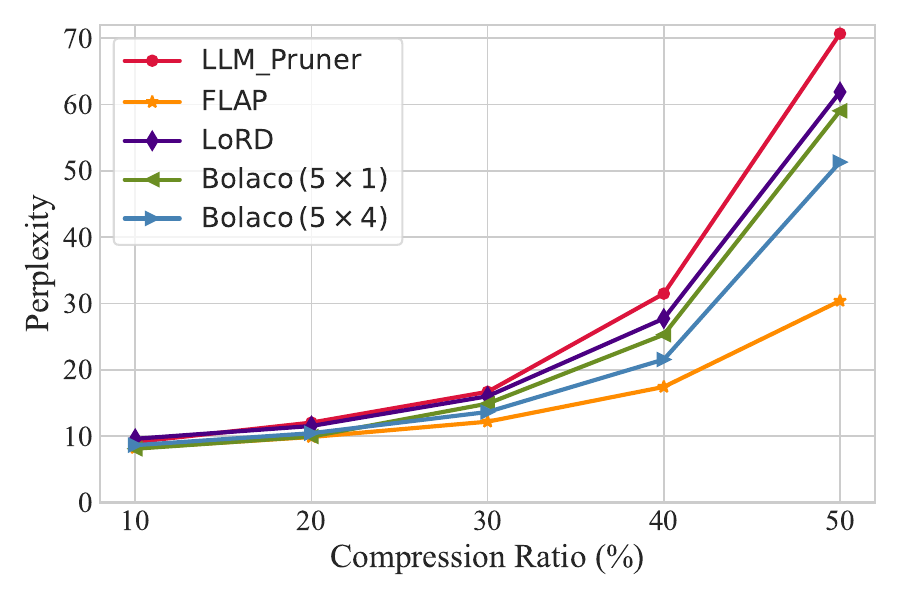} 
		\end{minipage}
		\label{fig:grid_4figs_1cap_4subcap_3}
	}
    	\subfigure[The perplexity of C4 on LLaMA-v2-13b]{
    		\begin{minipage}[b]{0.4\textwidth}
		 	\includegraphics[width=1\textwidth]{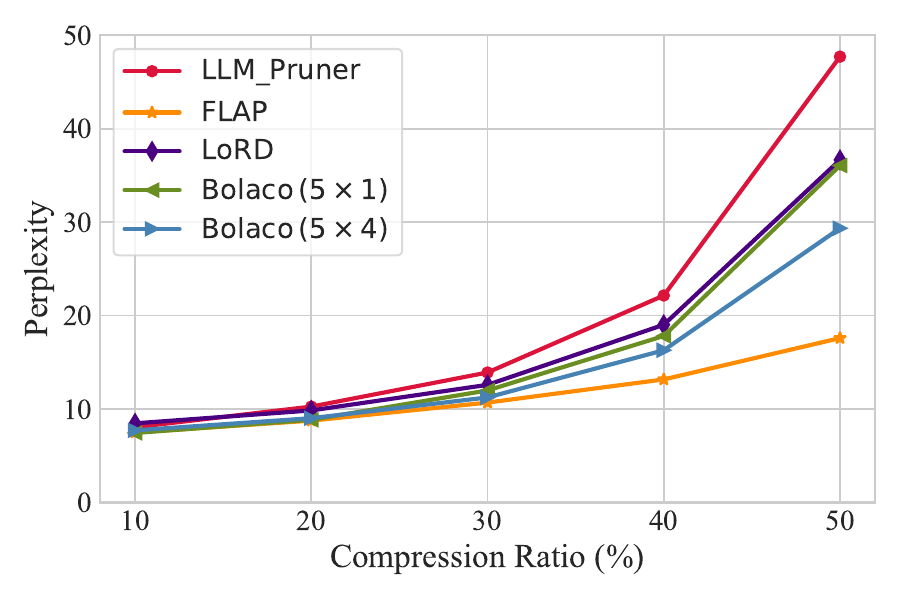}
    		\end{minipage}
		\label{fig:grid_4figs_1cap_4subcap_4}
    	}
        \\ 
	\subfigure[The perplexity of PTB on LLaMA-v2-7b]{
		\begin{minipage}[b]{0.4\textwidth}
			\includegraphics[width=1\textwidth]{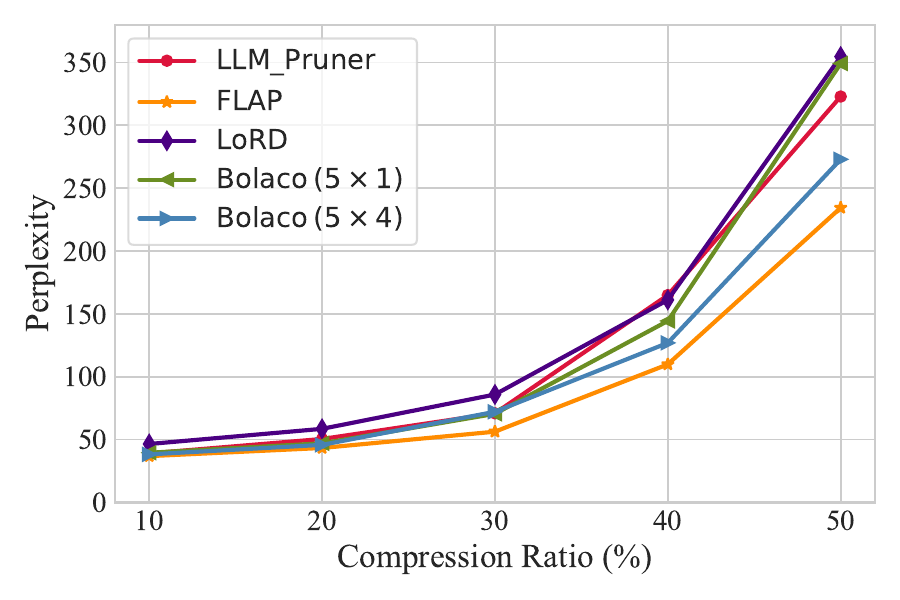} 
		\end{minipage}
		\label{fig:grid_4figs_1cap_4subcap_5}
	}
    	\subfigure[The perplexity of PTB on LLaMA-v2-13b]{
    		\begin{minipage}[b]{0.4\textwidth}
		 	\includegraphics[width=1\textwidth]{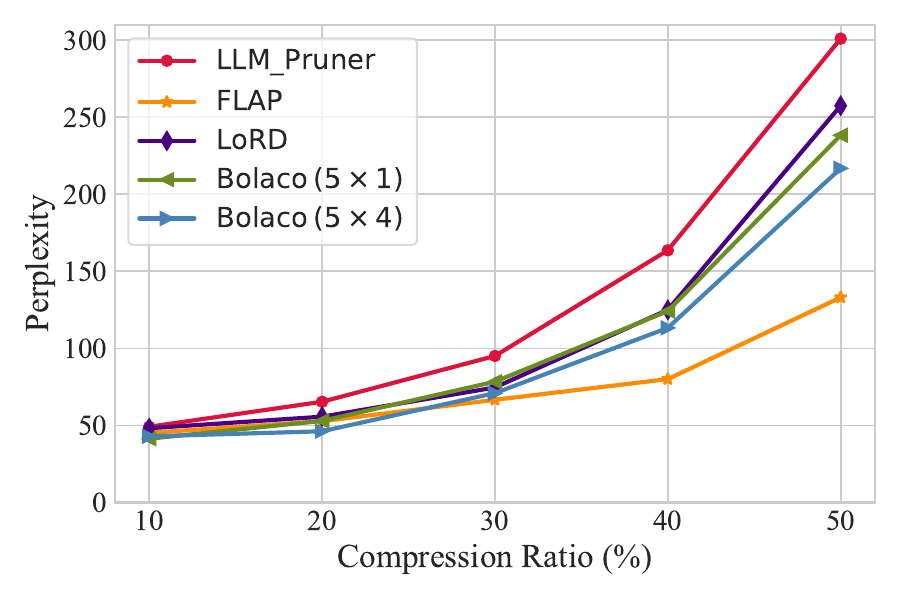}
    		\end{minipage}
		\label{fig:grid_4figs_1cap_4subcap_6}
    	}
	\caption{Language modeling capabilities at different compression ratios.}
	\label{fig:moreppl}
\end{figure*}

\input{table/efficient}
\input{table/allocation}
We report the statistic of original and compressed models in Table \ref{tab:efficient}, including the parameter count, MACs and memory requirements.
Statistical evaluation is conducted using the inference mode, where the model is fed a sentence consisting of 64 tokens.

To aid subsequent researchers in reproducing our results, Table \ref{tab:allocation} provides the low-rank allocations of Bolaco.
The elements of the array represent the low-rank dimensions for $attn\_q/k$, $attn\_o$, $mlp\_gate$, $mlp\_up$, and $mlp\_down$, respectively. `NA' denotes that the parameter is not compressed.

\section{Language Modeling Capabilities for Compressed Models}
Figure \ref{fig:moreppl} illustrates the perplexity changes on WikiText, PTB, and C4 datasets for different compression methods on LLaMA-v2-7b and 13b as the compression rate increases.

\section{Case Study}
\input{table/generate}
We showcase the generation results of the LLaMA-v2-7b and its compression model via Bolaco in Table \ref{tab:generate}.
We observe that models compressed via Bolaco tend to produce brief and repetitive responses to prompts without post-training. However, this issue can be resolved after efficient post-training, resulting in smooth and informative replies.

\subsection{The Generalization of Validation Data}
\input{table/generalization}
To verify the generalizability of the Bayesian optimization used in Bolaco across various validation data, we sample subsets from the Wikitext, C4, ArXiv, and Wikipedia pre-training datasets to serve as Bolaco’s validation data. 
Table \ref{tab:valid} presents the results of Bolaco on these validation data at 20\% compression ratio.
We observe that models optimized on different validation data exhibit different performance on a single test set, particularly in language modeling capabilities, likely due to the diverse linguistic features of the validation data. 
However, the average performance across multiple common sense reasoning datasets remains nearly identical, demonstrating the robustness of our method in general capabilities across different validation data.

\end{document}

%% file: article/1-introduction.tex
\section{Introduction}
In recent years, the emergence and application of large language models (LLMs) have served as a powerful stimulant for natural language processing and artificial intelligence \cite{chatgpt,OpenAI2023GPT4TR,bubeck2023sparks,yang2023dawn}.
Adhering to the scaling law \cite{kaplan2020scaling,hoffmann2022training}, researchers are continually seeking LLMs with more parameters and training data, aiming to achieve general models closer to human capabilities.
However, larger language models imply a larger overhead of computing resources. 
Therefore, when deploying LLMs, it is necessary to strike a balance between efficiency and performance \cite{wan2024efficient}. 
To achieve efficient LLMs, many compression techniques for LLMs are proposed, such as pruning \cite{pmlr-v202-frantar23a,sun2024a,ma2023llmpruner}, quantization \cite{frantar2023optq,lin2023awq,liu2023llmqat} and knowledge distillation \cite{gu2024minillm}.  

Among these methods, unstructured pruning and quantization can reduce the number of parameters or memory requirements by half or even more without significant performance degradation, but they require specialized GPU kernels to fully realize their acceleration potential.
In contrast, structured pruning can produce lightweight models that do not rely on specialized hardware.
Despite extensive research, the performance of structured pruning still lags significantly behind that of the original model. 
Low-rank compression (LRC) \cite{ben-noach-goldberg-2020-compressing,pmlr-v202-li23ap} is another promising compression technique. 
It decomposes the weight matrix into the product of two dense low-rank matrices, discarding unimportant parameter information during the decomposition process.
However, LRC remains under-explored in LLMs.

The keys to LRC are low-rank decomposition methods and low-rank dimension allocation.
Existing decomposition methods can generally be categorized into two types: weight-based and feature-based decomposition.
The former minimizes the reconstruction error of weight matrices by applying truncated SVD or weighted SVD \cite{ben-noach-goldberg-2020-compressing,hsu2022language,hua-etal-2022-numerical}. 
However, recent research \cite{NEURIPS2021_f56de5ef,Yu_Wu_2023} has discovered that the weights of most Transformer-based language models are typical of high rank or even close to full rank; thus, direct decomposition might result in significant error. 
In contrast, the model's features usually exhibit low-rank characteristics. 
Thus, more work focuses on the feature-based decomposition \cite{NEURIPS2021_f56de5ef,Yu_Wu_2023,kaushal2023lord}, which aims to minimize the reconstruction error of features. 
On the other hand, allocating suitable low-rank dimensions to different weight matrices according to the target compression ratio can also reduce the downside on the model's overall performance since they exhibit varying sensitivities to low-rank compression. 

When LRC is applied to LLMs, it encounters more new challenges.
First, it is challenging for LLMs to maintain their generality while achieving feature-based low-rank compression. 
This is because the feature space of LLMs is extremely high dimensional, making the feature distribution more complex, and the presence of outlier features may interfere with the accurate distribution estimation.
Thus, we utilize the pooled covariance matrix instead of the sample covariance matrix, which
enables a more accurate estimation of feature distributions \cite{Raninen_2022}.
Then, for low-rank dimension allocation, manual design struggles to achieve optimal results, and due to its vast search space, grid search requires a considerable amount of time.
We conduct empirical studies on the low-rank sensitivity of different types of parameters and observe significant variations among them.
Based on these findings, we categorize the parameters into groups, allowing each group to share the same low-rank dimensions. 
This approach effectively narrows down the search space, and furthermore, we utilize sample-efficient Bayesian optimization to determine the optimal low-rank allocation.
To evaluate the effectiveness of our proposed LRC method, we conduct experiments on two commonly used LLaMA-2 models \cite{touvron2023llama}.
Experimental results demonstrate
our proposed method can perform better than existing strong structured pruning and LRC methods in LLMs.
When combined with efficient post-training, our method obtains the latest state-of-art for the same settings, maintaining 98\% of the model's performance at the 20\% compression rate.

Overall, our main contributions include:
\begin{itemize}[leftmargin=*]
\setlength{\itemsep}{0pt}
\setlength{\parskip}{0pt}
\item We analyze the challenges that LLMs face in low-rank compression and demonstrate that LLMs represented by LLaMA exhibit vastly different sensitivities to low-rank compression across various parameters through empirical research. 
\item We propose a novel 
\textbf{B}ayesian \textbf{o}ptimization-based feature  \textbf{l}ow-r\textbf{a}nk \textbf{co}mpression  (\textbf{Bolaco}).
\item Extensive experiments show that our Bolaco outperforms the existing strong structured pruning and LRC methods in LLMs.
\end{itemize}

%% file: article/2-preliminary.tex
\section{Preliminary}
\begin{figure*}[t]
    \centering
\includegraphics[scale=0.55]{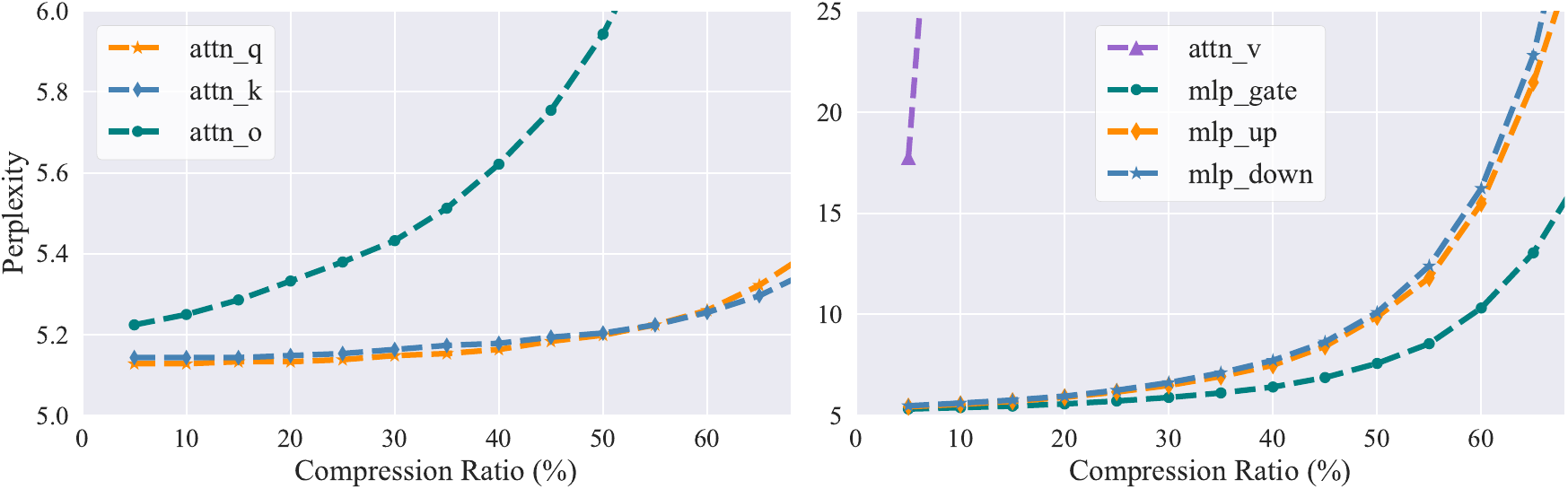}
    \caption{Sensitivity of different types of layers to low-rank compression. Each curve represents the compression of only that parameter type, with the horizontal axis indicating the compression ratio for that specific parameter type.}
    \label{fig:ppl}
\end{figure*}
In this section, we summarily introduce the foundation of low-rank factorization in model compression, and then empirically show that different layers of the Transformers-based generative large language model have different low-rank sensitivities.
\subsection{Weight-based and Feature-based Low-rank Decomposition}
The low-rank decomposition reduces the number of parameters by decomposing the linear layer weights into two low-rank matrices.
Weight-based factorization is one naive method.
For a linear layer $\boldsymbol{W} \in \mathbb{R}^{d_2 \times d_1}$, 
according to the Eckart–Young–Mirsky theorem, the trunced singular value decomposition (SVD) provides the optimal solution: $\boldsymbol{W}=\boldsymbol{U} \boldsymbol{\Sigma} \boldsymbol{V}^{T}, \boldsymbol{A}=\boldsymbol{V}_{r}^{T}, \boldsymbol{B}=\boldsymbol{U}_{r} \boldsymbol{\Sigma}_{r},$
where $\boldsymbol{A} \in \mathbb{R}^{r \times d_1}, \boldsymbol{B} \in \mathbb{R}^{d_2 \times r}$, $\boldsymbol{\Sigma}_r$ is the top-$r$ largest singular values, $\boldsymbol{U}_{r}$ and $\boldsymbol{V}_{r}$ are the corresponding singular vectors.
If $r < d_1d_2/(d_1+d_2)$, the factorization can reduce the total parameter amount.
However, in the vast majority of cases, the weights of PLMs have a high rank, and a direct truncated SVD decomposition on the weights would lead to significant reconstruction errors~\cite{NEURIPS2021_f56de5ef}. In comparison, the representation space of PLMs exhibits a clear low-rank property~\cite{Yu_Wu_2023}. Therefore, another line of work has considered feature-based factorization:
\begin{equation}
    \begin{aligned}
        & \underset{\boldsymbol{B}, \boldsymbol{A}}{\text{min}}  ||\boldsymbol{WX} - \boldsymbol{B}\boldsymbol{AX}||_{F} \ \\
& \text{s.t. } \text{rank}(\boldsymbol{B}\boldsymbol{A}) = r.
    \end{aligned}
    \label{eq:featue}
\end{equation}
For the linear layer $\boldsymbol{Y} = \boldsymbol{WX}$, \citet{NEURIPS2021_f56de5ef} obtain the optimal solution to Eq. \ref{eq:featue} by simultaneously performing the SVD decomposition of the weight and features.
\citet{Yu_Wu_2023} propose a more efficient Atomic Feature Mimicking (AFM) method, which utilizes the PCA decomposition to find the projection matrices:
\begin{equation}
    \begin{aligned}
         Cov(\boldsymbol{Y}) &= \boldsymbol{U \Sigma U}^T \\
         \boldsymbol{Y}-E[\boldsymbol{Y}] &= \boldsymbol{U}_{r}\boldsymbol{U}_{r}^{T}(\boldsymbol{WX}-E[\boldsymbol{Y}]),
    \end{aligned}
    \label{eq:afm}
\end{equation}
where $Cov(\boldsymbol{Y}) \in \mathbb{R}^{d_2 \times n}$, $E[\boldsymbol{Y}] \in \mathbb{R}^{d_2}$ is the covariance and mean of features.
Thus, the original linear layer can be replaced by $\boldsymbol{B}=\boldsymbol{U}_r \in \mathbb{R}^{d_2 \times r}$, $\boldsymbol{A}=\boldsymbol{U}_r^T\boldsymbol{W} \in \mathbb{R}^{r \times d_1}$ and the bias compensation $\boldsymbol{b}=(\boldsymbol{I}-\boldsymbol{U}_{r}\boldsymbol{U}_{r}^{T})E[\boldsymbol{Y}]$.
We have observed that the current mainstream LLMs also exhibit characteristics of high-rank weights and low-rank features. Therefore, in this paper, we focus on the feature-based low-rank factorization.

\subsection{Different Layers Exhibit Varying Degrees of Low-rank Sensitivity}
\label{preli}
Another challenge in LRC is allocating varying low-rank compression rates to different layers.
Previous works have empirically or theoretically demonstrated that different components of Transformer-based masked language models and visual models exhibit distinct low-rank properties, such as the features of the self-attention modules having a lower rank compared to those of the feed-forward modules~\cite{dong2023attention,anagnostidis2022signal}. 
These findings provide prior guidance for low-rank compression. However, detailed studies on current mainstream LLMs are still lacking.
Therefore, we take the LLaMA-v2-7b as an example to study the low-rank sensitivity within each layer across different types of layers and the same type of layers.
Llama-family LLMs have seven distinct parameter categories: \textit{attn\_q}, \textit{attn\_k}, \textit{attn\_v}, \textit{attn\_o}, \textit{mlp\_up}, \textit{mlp\_down}, and \textit{mlp\_gate}. We evaluate the perplexity changes on Wikitext-2 \cite{merity2016pointer} for each category under varying low-rank compression ratios.
As Figure \ref{fig:ppl} shows, at the same low-rank compression rate, distinct layers exhibit notable performance variations.
For \textit{attn\_q} and \textit{attn\_k}, they demonstrate robustness to low-rank compression, with an increase in perplexity not exceeding 2\% even at a compression rate of 60\%. In contrast, \textit{attn\_v}, with an equivalent parameter count, exhibits high sensitivity, leading to a significant surge in perplexity with compression rates even below 5\%.
Therefore, assigning the same low-rank compression rate to different types of layers during low-rank compression of LLM is a sub-optimal solution. 
In addition to the differences, we also observe certain similarities, e.g., \textit{attn\_q} and \textit{attn\_k} have similar low-rank sensitivities.
More empirical study results are shown in Appendix \ref{sec:sensitive}.

%% file: article/3-method.tex
\section{Methodology}

\subsection{Feature-Based Low-Rank Decomposition in High-Dimensional Spaces}
An efficient feature-based low-rank decomposition method performs PCA on features to identify the optimal low-rank matrices.
To achieve general task-agnostic compression, we follow the setup of prior work \cite{frantar-sparsegpt,sun2023wanda,ma2023llmpruner}, utilizing a subset of the pre-training data as calibration data $\mathcal{D}_{cal}=\{x_i\}_{i=1}^n$.
As described in Eq.\ref{eq:afm}, we first estimate the covariance matrix of the entire feature space distribution $\mathcal{Y}$ with the sample covariance matrix (SCM) of the calibration data features:
\begin{equation}
    Cov_{S}(\boldsymbol{Y})=\frac{1}{n-1}\sum^n_{i=1}(\boldsymbol{y}_i-\boldsymbol{\bar{y}})^T(\boldsymbol{y}_i-\boldsymbol{\bar{y}}),
\end{equation}
where $\boldsymbol{y}_i$ represents the feature of $x_i$, $\boldsymbol{\bar{y}}$ refers to the mean of all calibration data features.
However, LLMs typically have high-dimensional feature spaces (e.g., the intermediate size of LLaMA-v2-7b has exceeded 10,000 dimensions).
Precisely estimating the covariance matrix in such high-dimensional spaces has always been a statistical challenge, as the SCM does not effectively estimate the covariance of high-dimensional distributions.
For instance, calibration data sampled from pre-training datasets may introduce outlier features due to low-quality text or inadequate sampling. 
In high-dimensional spaces, these outlier features are difficult to identify due to the ``curse of dimensionality'', and their impact is further exacerbated in estimating high-dimensional covariance matrices due to the sparsity of data points.
Thus, to estimate the covariance of the feature space more robustly and accurately, we propose using the pooled covariance matrix (PCM) in place of the SCM. 
We split the calibration data into $m$ groups. 
For each group, we can calculate the SCM $Cov_S(\boldsymbol{Y_k})$, then the pooled covariance matrix is:
\begin{equation}
    Cov_P(\boldsymbol{Y})=\frac{1}{m}\sum_{k=1}^mCov_S(\boldsymbol{Y}_k)
\end{equation}

\subsection{Low-Rank Allocation Based on Bayesian Optimization}
\label{sec:bo}
\begin{figure*}[t]
    \centering
\includegraphics[scale=0.55]{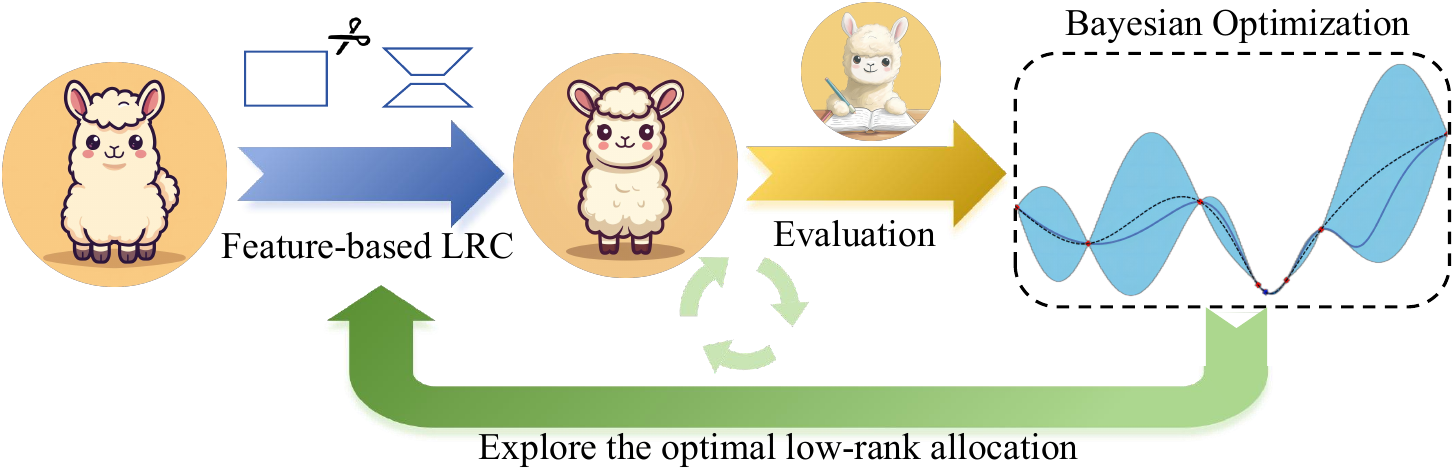}
    \caption{Illustration of our Bolaco. It initializes a low-rank dimension allocation and compresses the model via feature-based low-rank compression. Then, it evaluates the compression performance and optimizes the low-rank dimension allocation through Gaussian process-based Bayesian optimization.}
    \label{fig:bbo}
\end{figure*}
As investigated in Section \ref{preli}, different types of layers, and even each individual layer, exhibit varying sensitivities to low-rank compression. Therefore, allocating distinct compression ratios to different layers is crucial to achieve the desired compression rate with minimal performance degradation.
For a LLM $f(\cdot; \boldsymbol{\theta})$, we compress it with the set of low-rank compression ratios $\boldsymbol{\lambda}=\{\lambda_i\}_{i=1}^{k}$.
We use a task-agnostic evaluation dataset $\mathcal{D}$ to evaluate performance of the compressed model $f(\cdot; \boldsymbol{\theta},\boldsymbol{\lambda})$, such as the perplexity on a subset of pretraining data.
Therefore, the optimization objective of low-rank allocation can be formulated as: 
\begin{equation}
\begin{aligned}
    \min_{\lambda \in \mathcal{V}} H(\boldsymbol{\lambda}) &=\mathbb{E}_{(x,y)\sim \mathcal{D}} h(f(x; \boldsymbol{\theta},\boldsymbol{\lambda}), y) \\
    s.t. &\Sigma \boldsymbol{\lambda}\leq \rho,
\end{aligned}
\label{obj}
\end{equation}
where $h(\cdot,\cdot)$ is the evaluation metric, $\rho$ is model's overall compression ratio.
For LLMs, searching the optimal low-rank allocation is a challenging optimization problem. 
First, the impact of the low-rank count allocated to different layers on the performance of the compressed model is combinatorial, and optimizing any one component independently may lead to a locally optimal solution. 
Then, due to LLMs' vast number of parameters, evaluating $H(\boldsymbol{\lambda})$ is very time-consuming. 
Therefore, we leverage sample-efficient Bayesian optimization (BO) \cite{xu2022lower} to optimize Eq \ref{obj}.
BO estimates the objective $H(\boldsymbol{\lambda})$ with a stochastic surrogate model and updates the posterior estimation of $H(\boldsymbol{\lambda})$ based on the results of each search step.
We utilize the Gaussian process $\mathcal{N}(\mu(\cdot), \sigma^2(\cdot))$ as the surrogate model. 
Given the previous $t-1$ search steps $\{\boldsymbol{\lambda}_1,\cdots,\boldsymbol{\lambda}_{t-1}\}$ and their evaluation $H_{t-1}=[H(\boldsymbol{\lambda}_1),\cdots,H(\boldsymbol{\lambda}_{t-1})]$, the surrogate model is updated as:
\begin{equation}
    \begin{gathered}
        \mu(\boldsymbol{\lambda})=\boldsymbol{k}(\boldsymbol{K} + \eta^2 \boldsymbol{I})^{-1} H_{t-1} \\
        \sigma^2(\boldsymbol{\lambda}) = k(\boldsymbol{\lambda},\boldsymbol{\lambda}) - \boldsymbol{k}^T(\boldsymbol{K} + \eta^2 \boldsymbol{I})^{-1}\boldsymbol{k},
    \end{gathered}
\end{equation}
where $k(\cdot,\cdot)$ is a kernel function, 
$\boldsymbol{k}=(k(\boldsymbol{\lambda},\boldsymbol{\lambda}_i))_{i \in [t-1]}$,
$\boldsymbol{K}=(k(\boldsymbol{\lambda}_i,\boldsymbol{\lambda}_j))_{i,j \in [t-1]}$, and $\eta^2\boldsymbol{I}$ is the white kernel to model observation noise.

After obtaining the posterior estimation of $H(\boldsymbol{\lambda})$ (i.e., $H(\boldsymbol{\lambda}) \sim \mathcal{N}(\mu(\boldsymbol{\lambda}), \sigma^2(\boldsymbol{\lambda}))$), BO determines the next compression rate allocation state through the acquisition function.
Expected improvement (EI) is a popular and effective acquisition function:
\begin{equation}
    \begin{gathered}
    \alpha(\boldsymbol{\lambda})=\mathbb{E}_{H(\boldsymbol{\lambda})}\left[ \max \left\{ 0, H^{\prime} -H(\boldsymbol{\lambda})\right\} \right] \\
    \boldsymbol{\lambda}_t = \underset{\boldsymbol{\lambda}}{\mathrm{argmax}} \ \alpha(\boldsymbol{\lambda}),
    \end{gathered}
\end{equation}
where $H^{\prime}=\min_{i \in [t-1]} H(\boldsymbol{\lambda}_i)$, it means the minimal value observed so far.
Then, BO chooses the point with the greatest EI to explore.
After obtaining the optimal ratio $\boldsymbol{\lambda}^*$, we can determine the allocated rank: $r_i=(1-\lambda_i)d_1d_2/(d_1+d_2)$.
To fully leverage the acceleration effect of GPU matrix multiplication, we adhere to Nvidia's user guidelines\footnote{https://docs.nvidia.com/deeplearning/performance/dl-performance-matrix-multiplication/index.html\#gpu-imple} by rounding the low-rank dimensions to the nearest multiple of eight.

The evaluation metric and validation data play a significant role in the optimization performance of BO. They must meet two criteria: cost-effectiveness and accurately reflect actual changes in performance. To this end, we propose a sensitive-based sampling method. This method randomly samples $n$ allocation schemes, calculates the variance of the perplexity of each sample under different allocations, and selects the top-$k$ samples as validation data.
In addition, considering the smaller validation set may not comprehensively reflect the LLM's performance, potentially leading to over-fitting in the validation set.
To prevent BO from blindly improving the compressed model's language modeling performance on the validation set, we aim to make the compressed model have a prediction distribution for the next word close to the original model.
Hence, we employ the reverse KL divergence to quantify the difference:
\begin{equation}
    \mathcal{L}(\boldsymbol{\theta},\boldsymbol{\lambda})=D_{KL}(f(x;\theta)||f(x;\theta,\lambda)).
\end{equation}

\subsection{Post-training}
After low-rank compression, there remains a noticeable performance gap between the compressed model and the original LLM. 
To further bridge this gap, following \citet{ma2023llmpruner}, we perform efficient low-rank subspace post-training on the compressed model.
However, if we apply the original LoRA \cite{hu2022lora} to the low-rank compressed model, the tunable low-rank parameters may not be in the same subspace as the low-rank compressed model parameters, leading to an increase of the parameters' rank after merging.
Therefore, inspired by the ELoRA \cite{2024elora}, we select the subspace of compressed model parameters as fixed low-rank matrices and adjust the subspace by trainable vectors:
\begin{equation}
    \boldsymbol{Y}=(\boldsymbol{BA}+\boldsymbol{\Lambda_bB_{r^\prime}\Lambda_dA_{r^\prime}})\boldsymbol{X},
\end{equation}
where $\boldsymbol{B_{r^\prime}} \in \mathbb{R}^{d_2\times r^\prime}$ and $ \boldsymbol{A_{r^\prime}} \in \mathbb{R}^{r^\prime \times d_1}$ are fixed subspace of $\boldsymbol{B}$ and $\boldsymbol{A}$, $\boldsymbol{\Lambda_b}$ and $\boldsymbol{\Lambda_d}$ are diagonal matrices.
During the post-training, we only tune elements on the diagonal of $\boldsymbol{\Lambda_b}$ and $\boldsymbol{\Lambda_d}$.

%% file: article/4-experiments.tex
\section{Experiments}

\subsection{Baseline and Datasets}
We compare our method with the competitive structured pruning and low-rank compression methods in LLMs: LLM-Pruner \cite{ma2023llmpruner}, FLAP \cite{an2023fluctuationbased}, SliceGPT \cite{ashkboos2024slicegpt}, LoRD \cite{kaushal2023lord}, ASVD \cite{yuan2023asvd}. We provide the detailed description of baseline methods in Appendix \ref{sec:baseline}.

To evaluate the effectiveness of our proposed low-rank compression method in the task-agnostic setting, we conduct experiments in seven zero-shot common sense reasoning datasets: BoolQ \cite{clark-etal-2019-boolq}, PIQA \cite{piqa}, HellaSwag \cite{zellers-etal-2019-hellaswag}, WinoGrande \cite{10.1145/3474381}, ARC-easy/challenge \cite{clark2018think} and OpenbookQA \cite{mihaylov-etal-2018-suit}.
We also report the perplexity of the compressed model on the WikiText2 \cite{merity2016pointer}, PTB \cite{marcus-etal-1993-building}, and C4 \cite{JMLR:v21:20-074} datasets to evaluate its language modeling capabilities.

\subsection{Experimental Details}
In our main experiments, we apply our method to LLaMA-v2-7b and LLaMA-v2-13b.
We randomly select 1,024 samples from the training set of C4 as the calibration data. Each sample has a sequence length of 4,096.
To estimate the covariance matrix while saving memory usage, we employ the Welford's online algorithm \cite{welford}. For the pooled covariance matrix, we partition the calibrated data into 32 groups.
During the Bayesian optimization, we utilize the Matern kernel as the covariance function.
We randomly sample 20 low-rank allocation schemes and select the top-100 samples with greatest perplexity variance of Wikipedia as the evaluation data.
Each sample has a sequence length of 4,096 (4k tokens).
Considering that Bayesian optimization is not well-suited for high-dimensional scenarios, we conduct experiments with two settings based on the observations in Section \ref{preli}:
(a) $5\times1$: We allow $attn\_q$ and $attn\_k$ to share a low-rank dimension, and the same type of parameters across different layers to also share a low-rank dimension, thus BO only optimizes 5 parameters;
(b) $5\times4$: Building on the setup of (a), we divide the model's layers into 4 groups in sequence, with no parameter sharing between different groups, resulting in BO needing to optimize 20 parameters.
Moreover, given that the parameters of the FFN module are more sensitive to low-rank compression than those of the attention module, we set the length scale for the attention and FFN parameters in the Matern kernel to 1.0 and 0.8, respectively, to emphasize the more significant impact of FFN parameters' rank changes on model performance.
We run 50 epochs BO to search the optimal low-rank allocation.
At the post-training stage, following LLM-Pruner, we use the Alpaca dataset \cite{alpaca} and train 2 epochs.
More details can be found in the Appendix \ref{sec:details}.

\subsection{Main Results}
\begin{figure}[t]
    \centering
    \includegraphics[scale=0.52]{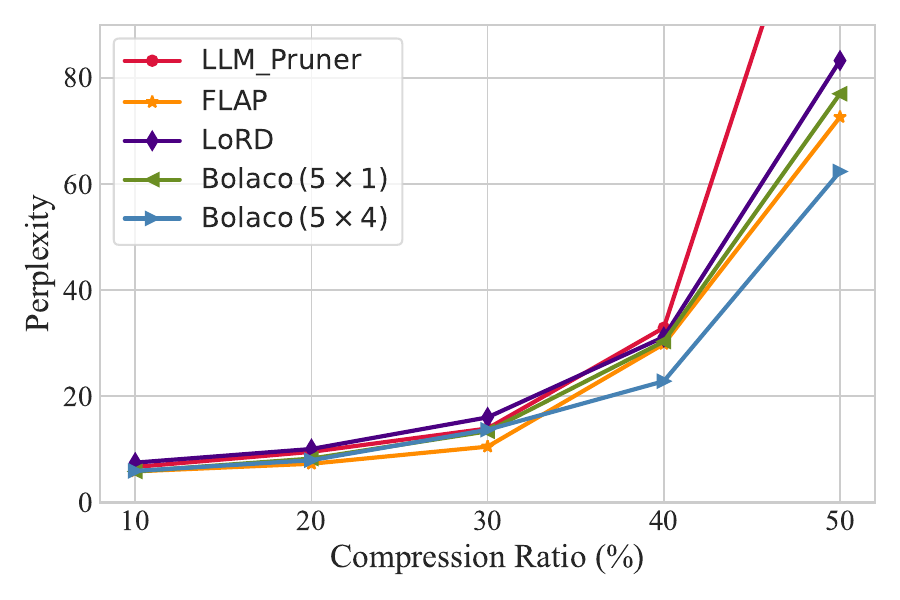}
    \caption{The perplexity of WikiText2 on LLaMA 2-7b with different compression ratios.}
    \label{fig:wiki_ppl}
\end{figure}
\input{table/new_result}
We report the perplexity of language modeling for various compression methods at different compression ratios in Figure \ref{fig:wiki_ppl} and \ref{fig:moreppl}, and the zero-shot common sense reasoning results in Table \ref{tab:zeroshot} and \ref{tab:llama13b-zeroshot} (in Appendix).
In terms of language modeling capabilities, FLAP demonstrates strong competitiveness, particularly when the compression rate exceeds 30\%, where FLAP's perplexity is slightly better than our Bolaco ($5\times1$). However, in the 7b model, Bolaco ($5\times4$) achieves the best language modeling performance at high compression rates. Nevertheless, in the 13b model, despite Bolaco ($5\times4$) still leading other compression techniques, it maintains a certain gap from FLAP.
For zero-shot tasks, our method significantly outperforms all baselines without any further post-training, achieving an average performance increase of 1.5-2\% across seven datasets.
After post-training with only about 1\textperthousand~parameters and 3 hours, our method further narrows down the performance difference between the compressed model and the original model.
It retains 96\%-98\% of the original model's performance at the 20\% compression ratio, and at a 30\% compression ratio, it maintains 91\%-95\% of the performance.
Comparing the $5\times1$ and $5\times4$ setting, we find that the performance difference between the two is not significant. 
At the 20\% compression ratio, simply allocating different low-rank dimensions to different types of parameters suffices to achieve the best current performance. 
However, at the 30\% compression rate, the $5\times4$ setting outperforms the $5\times1$, indicating that more granular low-rank assignments contribute to enhanced performance in compressed models at higher compression rates.

%% file: table/new_result.tex
\begin{table*}[t]
\centering
\resizebox{\textwidth}{!}{
\begin{tabular}{ll|ccccccc|c}
\toprule
Ratio& Methods& BoolQ& PIQA& HellaSwag& WinoGrande& ARC-e& ARC-c& OBQA& Average\\
\hline
0\%& LLaMA-v2-7b& 77.74& 78.07& 75.97& 68.98& 76.30& 46.33& 44.20& 66.80\\
\hline
\multirow{10}*{20\%}& LLM-Pruner& 63.27& 76.12& 67.93& 64.80& 68.73& 38.65& 40.00& 59.93\\
& LLM-Pruner (w/ PT) & 66.45 & 76.28 & 70.90 & 65.75 & 70.62 & 39.59 & 43.20 & 61.83 \\
& FLAP & 70.21 & 75.24 & 69.34 & 66.30 & 67.30 & 39.42 & 37.40 & 60.74 \\
& SliceGPT & 46.73 & 69.04 & 58.98 & 64.33 & 60.31 & 35.07 & 40.40 & 53.55 \\

& LoRD & 72.60 & 73.56 & 63.70 & 65.90 & 69.70 & 37.71 & 39.20 & 60.34\\
& ASVD & 73.61 & 71.93 & 66.05 & 64.17 & 65.24 & 36.26 & 37.40 & 59.24 \\

& Bolaco ($5\times1$)& 72.17& 75.52& 66.76& \underline{67.72}& \underline{73.02}& 38.74& 40.60& 62.08 \\
& Bolaco ($5\times1$ w/ PT)& 73.79 & \textbf{77.53} & \textbf{72.72} & \textbf{68.11} & \textbf{73.19} & \textbf{42.24} & \underline{43.60} & \textbf{64.45} \\

& Bolaco ($5\times4$)& \underline{75.05}& 75.46& 67.12& 67.01& 72.05& 38.91& 42.40 & 62.57\\
& Bolaco ($5\times4$ w/ PT) & \textbf{75.84} & \underline{76.61} & \underline{71.70} & 65.67 & 72.60 & \underline{41.81} & \textbf{45.00} & \underline{64.18} \\
\hline

\multirow{10}*{30\%}& LLM-Pruner& 52.51 & 71.93 & 59.49 & 58.72 & 61.41 & 33.96 & 36.60 & 53.52 \\
& LLM-Pruner (w/ PT) & 63.30 & \textbf{76.01} & 65.23 & 64.25 & 66.62 & 37.20 & 40.20 & 58.97 \\

& FLAP & 66.88 & 72.74 & 63.80 & 64.01 & 60.65 & 34.47 & 36.40 & 56.99 \\
& SliceGPT & 39.11 & 63.38 & 49.16 & 62.47 & 55.72 & 31.48 & 32.80 & 47.73 \\
& LoRD & 69.63 & 70.46 & 55.87 & 64.17 & 63.80 & 32.59 & 35.00 & 55.93 \\
& ASVD & 59.42 & 55.93 & 35.05 & 52.25 & 34.30 & 26.45 & 26.60 & 41.43\\

& Bolaco ($5\times1$)& 68.26 & 72.09 & 57.46 & \textbf{65.87} & 65.19 & 32.85 & 37.20 & 56.99\\
& Bolaco ($5\times1$ w/ PT)& 70.34 & 74.32 & \underline{67.81} & 65.04 & \underline{69.02} & \underline{38.31} & \underline{41.80} & \underline{60.95} \\

& Bolaco ($5\times4$)& \underline{70.37} & 71.44 & 59.62 & 64.80 & 66.46 & 34.39 & 38.60 & 57.95 \\
& Bolaco ($5\times4$ w/ PT)& \textbf{71.83} & \underline{75.19} & \textbf{68.03} & \underline{65.67} & \textbf{69.15} & \textbf{38.74} & \textbf{42.40} & \textbf{61.57} \\
\bottomrule
\end{tabular}}
\caption{Zero-shot performance of the  compressed LLaMA-v2-7b models. w/ PT means the method with post-training. \textbf{Bold} denotes the best result at the same compression ratio, while \underline{underline} indicates the second best result. }
\label{tab:zeroshot}
\end{table*}

%% file: article/5-results.tex
\section{Analysis and Discussion}

\subsection{Impact of Calibration Data and Covariance Estimation}
\input{table/cov_estimate}
Accurate estimation of feature distribution is crucial for the feature-based low-rank decomposition, which primarily depends on the number of calibration samples and the accuracy of the covariance matrix estimation.
Thus, we investigate the impact of the two factors on LLaMA-v2-7b at the 20\% compression ratio.
In this experiment, we do not account for the effects of low-rank dimensions allocation, and maintain consistency with the settings of LoRD.
As results shown in  Table \ref{tab:calib}, as the calibration dataset size gradually increases, we observe a consistent improvement in both the language modeling capabilities and the performance on downstream tasks of the compressed model. 
Therefore, given sufficient data and computational resources, expanding the calibration dataset is a reliable method for enhancing the performance of compressed models.
On the other hand, comparing the two covariance estimation methods, there is no significant difference in their language modeling capabilities. 
However, for downstream common sense reasoning tasks, the pooled SCM achieves an average improvement of 0.3 points across seven datasets without any additional burden.

\subsection{Impact of Objective Function}
\input{table/objective}
We explore the impact of the objective function in the Bayesian optimization stage.
We conduct experiments on LLaMA-v2-7b and report results in Table \ref{tab:obj}.
Overall, incorporating the reverse KL divergence (RKL) between the compressed model and the original model's predictive distribution into the objective function can lead to a better low-rank dimensions allocation.
Especially in the $5\times4$ setting, which is more difficult to optimize for Bayesian optimization, the performance gains from RKL term are even more obvious.
We suppose that the RKL term may serve two roles. Firstly, as a regularization term, it prevents overfitting on smaller validation sets during BO. 
Although the compressed model exhibits a slight increase in perplexity on the language modeling dataset at the 20\% compression rate with the $5\times4$ setting, there is a significant improvement in performance on downstream tasks. 
Secondly, incorporating the RKL term may smooth the objective function, enabling the Gaussian process surrogate model to more accurately approximate the real black-box objective function.

\subsection{The Transferability of Rank Allocation}
\begin{figure}[t]
    \centering
\includegraphics[scale=0.55]{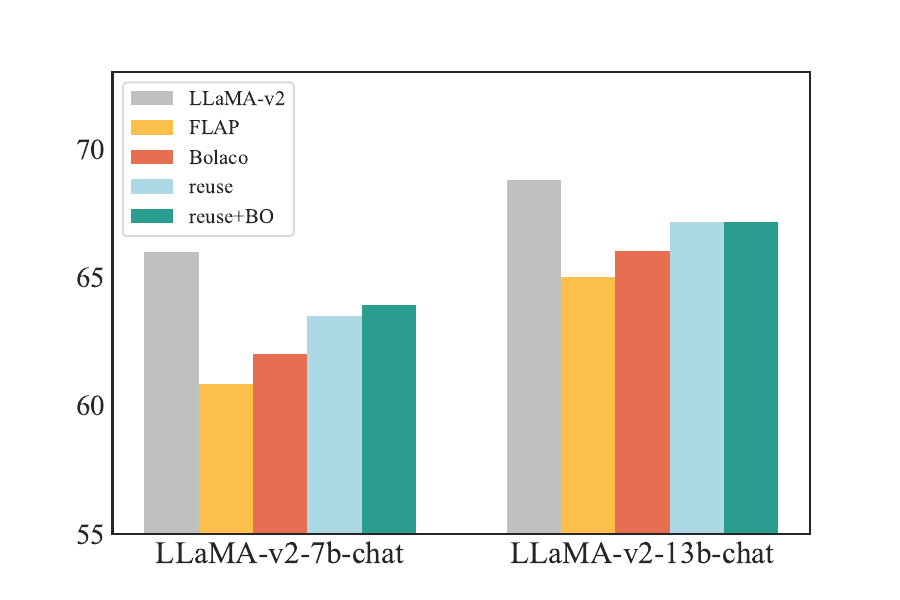}
    \caption{The average performance on zero-shot tasks about the transferability of rank allocation.}
    \label{fig:transfer}
\end{figure}
In practical applications, we may utilize a variety of fine-tuned models based on the LLaMA foundation model.
If we perform Bayesian optimization from scratch to optimize the low-rank allocation for each model, it will waste a significant amount of time and computational resources. 
Hence, we investigate whether the low-rank allocation of the base model can be transferred to the corresponding fine-tuned models.
We transfer the allocation of LLaMA-v2-7b/13b to LLaMA-v2-7b/13b-chat, respectively.
We consider two migration strategies: \textbf{a)} directly reusing the low-rank allocation of the base model and \textbf{b)} using the low-rank allocation of the base model as the initial value for Bayesian optimization and then optimizing only 20 epochs.
As Figure \ref{fig:transfer} shows, direct reusing can achieve results that outperform all baseline methods, even the Bayesian optimization from scratch. 
If 20 epochs of Bayesian optimization follow reuse, there is a chance to find an even better low-rank allocation.

\subsection{The Effectiveness of Validation Data Sampling}
Table \ref{tab:select} shows results on LLaMA-v2-7b at 20\% compression ratios under Wikipedia and its sampled data.
The top-100 and bottom-100 represent the 100 samples with the highest and lowest perplexity variances, respectively.
BO can optimize a good result when using Wikipedia and the top-100 sampled data for validation, showing that our method can sample a smaller subset for improving validation efficiency while maintaining performance comparable to the entire dataset.
Conversely, with the bottom-100 sampled data, BO's optimization performance is significantly inferior, with performance similar to unoptimized LoRD.
Furthermore, we observe that the top-100 samples (6.09 ppl) have higher perplexity than the bottom-100 samples (3.25 ppl) in the original LLaMA-v2-7b, indicating that the bottom-100 samples are already well-modeled and are very robust to model compression. This data may not truly reflect the performance change caused by model compression.
Therefore, we suggest selecting validation data that is more sensitive to compression, typically samples with slightly worse language modeling performance. Similar to the ``buckets effect'', these samples may represent the performance boundaries of LLMs. Considering the performance of these samples in the optimization process can maximize the overall performance of the compressed model.

%% file: table/cov_estimate.tex
\begin{table}[t]
\centering
\resizebox{0.48\textwidth}{!}{
\begin{tabular}{lcccc}
\toprule
 & Wikitext ($\downarrow$) & PTB ($\downarrow$) & C4 ($\downarrow$) & ZS ($\uparrow$) \\
\hline
\textit{Covariance estimate} \\
Naive SCM & 9.96 & 54.69 & 11.46 & 60.34\\
Pooled SCM & 9.93 & 54.68 & 11.45 & 60.64 \\
\hline
\textit{\# Samples} \\
128 & 10.55 & 56.29 & 11.99 & 60.26 \\
256 & 10.24 & 55.42 & 11.88 & 60.16 \\
512 & 10.30 & 55.03 & 11.61 & 60.56 \\
1,024 & 9.93 & 54.68 & 11.45 & 60.64 \\
\bottomrule
\end{tabular}
}
\caption{Impact of different covariance estimation methods and the number of calibration data. ``ZS'' denotes the average performance on seven zero-shot common sense reasoning datasets.}
\label{tab:calib}
\end{table}

%% file: table/objective.tex
\begin{table}[t]
\centering
\resizebox{0.48\textwidth}{!}{
\begin{tabular}{lcccc}
\toprule
&Wikitext ($\downarrow$) & PTB ($\downarrow$) & Zero-shot ($\uparrow$) \\
\hline
$20\%$ \\
PPL ($5\times1$) & 8.36 & 48.42 & 61.70\\
w/ RKL ($5\times1$) & 8.27 & 47.06 & 62.08\\
PPL ($5\times4$) & 8.07 & 47.96 & 60.98\\
w/ RKL ($5\times4$) & \textbf{7.96} & \textbf{45.84} & \textbf{62.57}\\
\hline
$30\%$ \\
PPL ($5\times1$) & 13.78 & 71.50 & 56.97\\
w/ RKL ($5\times1$)& 13.41 & 70.52 & 56.99\\
PPL ($5\times4$) & \textbf{12.65} & \textbf{68.85} & 57.57\\
w/ RKL ($5\times4$)& 13.70 & 72.14 & \textbf{57.95}\\
\bottomrule
\end{tabular}
}
\caption{Results under different objective function. }
\label{tab:obj}
\end{table}

%% file: article/6-related_work.tex
\section{Related work}

A common technique for low-rank factorization is 
SVD, which retains only the top-$r$ largest singular values and their corresponding singular vectors to obtain two rank-$r$ matrices.
\citet{ben-noach-goldberg-2020-compressing} first combine SVD with knowledge distillation, applying it to compress BERT.
Directly applying SVD decomposition implies an assumption that each parameter in the weight matrix equally affects the model performance. This contradicts many previous research, therefore, FWSVD \cite{hsu2022language} and TFWSVD \cite{hua-etal-2022-numerical} consider weighting the weight matrix using Fisher information.
\citet{NEURIPS2021_f56de5ef} observe that PLMs' weights are not inherently low-rank matrices. Therefore, directly applying SVD will result in significant reconstruction loss. However, they find that the product of data representation and weights is low-rank. Hence, they perform a global low-rank decomposition on it.
Following this observation, \citet{Yu_Wu_2023} propose the atomic feature mimicking (AFM) method to decompose the output features.
\citet{Ren2023LowRankPF} also observe the high rank phenomenon of PLM weights. They utilize iterative first-order unstructured pruning to reduce the rank of the weight matrix, and then apply Fisher information-weighted SVD decomposition for low-rank compression.
For LLMs, low-rank compression has not yet received the attention it deserves.
LoRD \cite{kaushal2023lord} applies AFM to code LLMs, demonstrating the potential of low-rank decomposition in compressing LLM.
Recently, \citet{sharma2023truth} conduct an in-depth study on the weight decomposition of LLMs and discover that the low-rank components of the weights encapsulate low-frequency information. By meticulously selecting low-rank components, it is possible to eliminate interfering signals and further improve LLMs' performance. However, their research does not propose a practical low-rank compression algorithm.

\input{table/select_data}

%% file: table/select_data.tex
\begin{table}[t]
\centering
\resizebox{0.48\textwidth}{!}{
\begin{tabular}{lcccc}
\toprule
&Wikitext ($\downarrow$) & PTB ($\downarrow$) & Zero-shot ($\uparrow$) \\
\hline
Wikipedia  & 7.98 & 46.85 & 62.27\\
Top-100  & \textbf{7.96} & \textbf{45.84} & \textbf{62.57} \\
Bottom-100 & 8.38 & 50.41 & 60.90 \\
\bottomrule
\end{tabular}
}
\caption{Results under different validation data.}
\label{tab:select}
\end{table}

%% file: article/7-conclusion.tex
\section{Conclusion}
In this paper, we attempt to unearth the potential of low-rank compression for lightweight universal LLMs.
We thoroughly investigate the challenges of low-rank compression in LLMs and the low-rank characteristics of features within LLMs. 
We propose a Bayesian optimization-based feature low-rank compression to address these challenges, incorporating pooled covariance estimation and Bayesian optimization for more precise feature distribution estimation and low-rank dimension allocation, respectively. Experimental results on the LLaMA 2 model demonstrate that our method significantly outperforms existing structured pruning and other low-rank compression techniques.

%% file: table/add_result.tex
\begin{table*}[t]
\centering
\resizebox{\textwidth}{!}{
\begin{tabular}{ll|ccccccc|c}
\toprule
Ratio& Methods& BoolQ& PIQA& HellaSwag& WinoGrande& ARC-e& ARC-c& OBQA& Average\\
\hline
0\%& LLaMA-v2-13b& 80.52& 79.05& 79.38& 72.14& 79.42& 49.23& 45.20& 69.27\\
\hline
\multirow{7}*{20\%}& LLM-Pruner& 66.33 & \underline{78.18} & 74.47 & 64.48 & 72.26 & \underline{45.90} & 44.20 & 63.69\\
& LLM-Pruner (w/ PT) & 67.06 & \textbf{78.94} & \underline{75.92} & 67.32 & 72.69 & 44.28 & \underline{44.60} & 64.40 \\

& FLAP & 71.28 & 76.55 & 74.67 & 69.53 & 72.56 & 44.03 & 42.00 & 64.37\\
& SliceGPT & 45.44 & 71.00 & 62.86 & 68.35 & 71.09 & 41.72 & 41.20 & 57.38 \\
& ASVD & 79.36 & 76.61 & 72.82 & 69.69 & 74.54 & 43.00 & 44.60 & 65.80\\
& LoRD & 78.47 & 76.01 & 69.58 & 71.03 & 74.33 & 40.87 & 44.40 & 64.96\\

& Bolaco ($5\times1$)& 80.00 & 76.50 & 73.25 & 70.24 & \underline{76.18} & 43.86 & \textbf{45.20} & 66.46\\
& Bolaco ($5\times1$ w/ PT)& \textbf{81.22} & 77.69 & \textbf{76.66} & \textbf{71.59} & \textbf{77.31} & \textbf{46.93} & 44.00 & \textbf{67.91} \\
& Bolaco ($5\times4$)& 80.58 & 76.22 & 71.44 & \underline{71.19} & 75.38 & 42.49 & 44.00 & 65.90 \\
& Bolaco ($5\times4$ w/ PT)& \underline{80.95} & 77.64 & 75.84 & 69.93 & 75.25 & 45.14 & 44.20 & \underline{67.00} \\
\hline
\multirow{7}*{30\%}& LLM-Pruner& 62.45 & 75.90 & 67.90 & 60.22 & 65.45 & 40.36 & \underline{44.60} & 59.55\\
& LLM-Pruner (w/ PT) & 68.29 & \textbf{76.66} & 72.03 & 64.09 & 69.20 & 41.13 & \textbf{45.40} & 62.40\\

& FLAP& 65.54& 74.81& 70.29& 67.48& 67.38& 38.23& 40.00& 60.53\\
& SliceGPT & 38.84 & 64.47 & 52.34 & 65.51 & 59.51 & 36.86 & 39.20 & 50.96 \\
& ASVD & 70.34 & 68.01 & 53.41 & 60.93 & 59.72 & 32.00 & 36.60 & 54.43\\
& LoRD & 75.05 & 73.88 & 63.08 & \textbf{69.46} & 69.78 & 39.16 & 38.60 & 61.29\\

& Bolaco ($5\times1$)& 79.20 & 74.97 & 65.23 & 67.32 & 72.35 & 39.25 & 41.20 & 62.79\\
& Bolaco ($5\times1$ w/ PT)& 78.78 & \underline{76.17} & \underline{73.04} & 68.51 & \textbf{74.75} & \underline{43.60} & 44.00 & \underline{65.55} \\
& Bolaco ($5\times4$)& \underline{80.24} & 74.48 & 66.77 & \underline{69.14} & 72.18 & 41.13 & 41.00 & 63.56 \\
& Bolaco ($5\times4$ w/ PT)& \textbf{80.40} & \textbf{76.66} & \textbf{73.42} & 69.06 & \underline{73.74} & \textbf{45.14} & 43.40 & \textbf{65.97} \\
\bottomrule
\end{tabular}}
\caption{Zero-shot performance of the  compressed LLaMA-v2-13b models. w/ PT means the method with post-training. \textbf{Bold} denotes the best result at the same compression ratio, while \underline{underline} indicates the second best result. }
\label{tab:llama13b-zeroshot}
\end{table*}

%% file: table/mistral.tex
\begin{table*}[t]
\centering
\resizebox{\textwidth}{!}{
\begin{tabular}{ll|ccccccc|c}
\toprule
Ratio& Methods& BoolQ& PIQA& HellaSwag& WinoGrande& ARC-e& ARC-c& OBQA& Average\\
\hline
0\%& Mistral-7B-v0.1 & 83.67 & 80.52 & 81.03 & 73.80 & 80.85 & 54.01 & 43.8 & 71.10\\
\hline
\multirow{5}*{20\%} & LLM-Pruner & 70.06 & \textbf{77.31} & \textbf{72.50} & 68.35 & 69.11 & 38.23 & \textbf{41.80} & 62.48\\
& LORD & 73.82 & 74.86 & 65.53 & 69.22 & 71.55 & 41.13 & 36.20 & 61.76\\
& Bolaco ($5\times1$) & \underline{74.13} & 76.01 & 66.26 & \underline{69.69} & \underline{74.24} & \underline{42.15} & \underline{39.40} & \underline{63.13}\\
& Bolaco ($5\times4$) & \textbf{77.58} & \underline{76.12} & \underline{67.44} & \textbf{70.09} & \textbf{74.96} & \textbf{42.41} & \underline{39.40} & \textbf{64.00}\\
\bottomrule
\end{tabular}}
\caption{Zero-shot performance of the compressed Mistral-7B-v0.1 models. \textbf{Bold} denotes the best result at the same compression ratio, while \underline{underline} indicates the second best result. }
\label{tab:mistral-zeroshot}
\end{table*}

%% file: table/efficient.tex
\begin{table}[t]
\centering
\resizebox{\textwidth/2}{!}{
\begin{tabular}{l|c|ccc}
\toprule
Method&Ratio&\#Params  & MACs & Memory \\
\hline
LLaMA 2-7b & 0\% & 6.74B & 423.98G & 12.62GiB\\
\hline
LLM-Pruner & 20\% & 5.42B & 340.48G & 10.16GiB \\
FLAP & 20\% & 5.45B & 342.30G & 10.22GiB \\
LoRD & 20\% & 5.45B& 370.12G& 10.32GiB \\
Bolaco ($5\times1$) & 20\% & 5.44B & 388.95G & 10.28GiB  \\
Bolaco ($5\times4$) & 20\% & 5.44B & 391.18G & 10.25GiB  \\
\hline
LLM-Pruner & 30\% & 4.84B & 302.83G & 9.17GiB \\
FLAP & 30\% & 4.80B & 300.72G & 9.04GiB \\
LoRD & 30\% & 4.79B & 341.91G & 9.07GiB \\
Bolaco ($5\times1$) & 30\% & 4.79B & 359.48G & 9.04GiB  \\
Bolaco ($5\times4$) & 30\% & 4.80B & 356.03G & 9.06GiB \\
\hline\hline
LLaMA 2-13b & 0\% & 13.02B & 824.26G & 24.45GiB \\
\hline
LLM-Pruner & 20\% & 10.48B & 662.95G & 19.75GiB \\
FLAP & 20\% & 10.48B & 663.85G & 19.64GiB \\
LoRD & 20\% & 10.49B & 717.86G & 19.79GiB \\
Bolaco ($5\times1$) & 20\% & 10.48B & 777.58G & 19.71GiB \\
Bolaco ($5\times4$) & 20\% & 10.48B & 772.16G & 19.69GiB \\
\hline
LLM-Pruner & 30\% & 9.21B & 581.40G & 17.35GiB \\
FLAP & 30\% & 9.21B & 582.72G & 17.29GiB \\
LoRD & 30\% & 9.21B & 663.15G & 17.38GiB \\
Bolaco ($5\times1$) & 30\% & 9.21B & 708.16G & 17.36GiB \\
Bolaco ($5\times4$) & 30\% &  9.21B & 694.58G & 17.35GiB\\
\bottomrule
\end{tabular}
}
\caption{Statistics of the compressed model.}
\label{tab:efficient}
\end{table}

%% file: table/allocation.tex
\begin{table}[!h]
\centering
\resizebox{\textwidth/2}{!}{
\begin{tabular}{l|l|c|c}
\toprule
Model & Method & Ratio & Low rank allocation \\

\hline
\multirow{10}*{LLaMA-v2-7b} & Bolaco ($5\times1$) & 20\% & [744, 1616, 2512, 2408, NA] \\
\cline{2-4}
 & \multirow{4}*{Bolaco ($5\times4$)} & \multirow{4}*{20\%} & [[680, 1728, 2960, NA, NA], \\
 & & & [968, 1888, 2536, 2640, 2632], \\
 & & & [408, 1488, NA, 2272, 2864], \\
 & & & [656, 496, 2824, 2448, 2280]] \\
\cline{2-4}
 & Bolaco ($5\times1$) & 30\% & [656, 1392, 2128, 2352, 2312] \\
\cline{2-4}
 & \multirow{4}*{Bolaco ($5\times4$)} & \multirow{4}*{30\%} & [[1016, 1632, 2376, 2384, 2384],  \\
& & & [840, 1632, 2384, 2376, 2384],  \\
& & & [408, 992, 2376, 2384, 2384],  \\
& & & [408, 560, 2384, 1896, 1792]] \\
\hline\hline

\multirow{10}*{LLaMA-v2-13b} & Bolaco ($5\times1$) & 20\% & [696, 1920, 2304, NA, 2504] \\
\cline{2-4}
 & \multirow{4}*{Bolaco ($5\times4$)} & \multirow{4}*{20\%} & [[792, 1696, 2864, 2880, 2976],\\
 & & & [944, 1440, 2512, 2296, 2920],\\
 & & & [656, 1112, 2496, 2480, 2912],\\
 & & & [1312, 904, 2264, NA, 1960]]\\

\cline{2-4}
& Bolaco ($5\times1$) & 30\% & [512, 1264, 2384, 2328, 2304] \\
\cline{2-4}
& \multirow{4}*{Bolaco ($5\times4$)} & \multirow{4}*{30\%} & [[528, 1536, 2384, 2376, 2384], \\
& & &  [1232, 1624, 2376, 2352, 2344], \\
& & &  [800, 1624, 2064, 2368, 2344], \\
& & &  [408, 408, 2352, 1936, 1680]]\\

\bottomrule
\end{tabular}
}
\caption{The low-rank allocation of our Bolaco.}
\label{tab:allocation}
\end{table}

%% file: table/generate.tex
\begin{table*}[!t]
\centering
\resizebox{\textwidth}{!}{
\begin{tabular}{l|l}
\toprule
Model & Generated Sentences \\
\hline
\multirow{8}*{LLaMA 2-7b} & Instruction: I sit in front of a computer all day. How do I manage and mitigate eye strain?\\
& Response: Eye strain is a common complaint among computer users. It is caused by the eye’s inability to focus on the computer screen.\\
&This is due to the fact that the computer screen is too close to the eyes. The eyes are also strained by the glare from the computer screen.\\
&The best way to manage and mitigate eye strain is to take frequent breaks from the computer. This will allow the eyes to rest and relax.\\
\cmidrule(r){2-2}
& Instruction: Translate the sentence ’I have no mouth but I must scream’ into Spanish.\\
& Response: No tengo boca pero debo gritar.\\
\cmidrule(r){2-2}
&Instruction: Who is Donald Trump? \\
& Response: Donald Trump is the 45th President of the United States.\\
\hline
\multirow{8}*{\shortstack{Bolaco (20\%) \\ (w/o tune)}} & Instruction: I sit in front of a computer all day. How do I manage and mitigate eye strain?\\
& Response: Most of the eye strain in this population is from staring at a computer monitor, which does not affect eye strain nearly as much \\
&as looking at a printed page. I recommend that people sit with their computer at arm’s length, in a room that has adequate lighting, and not \\
&looking at a monitor for prolonged periods. If the screen is too bright, it causes glare and eye strain. If the room is bright, it also causes eye \\
&strain. The eyes don’t work well with bright light.\\
\cmidrule(r){2-2}
& Instruction: Translate the sentence ’I have no mouth but I must scream’ into Spanish.\\
& Response: Mi boca no has a voz y tengo que scream.\\
\cmidrule(r){2-2}
&Instruction: Who is Donald Trump? \\
& Response: Trump is the president of the United States.\\
\hline
\multirow{12}*{\shortstack{Bolaco (20\%) \\ (w/ tune)}} & Instruction: I sit in front of a computer all day. How do I manage and mitigate eye strain?\\
& Response: Here are some tips to help you manage and mitigate eye strain: \\

&1. Position your monitor correctly: Make sure that your monitor is at least 20-25 inches away from you. If it’s too close, your eyes will have to\\
&strain to focus on it, which can lead to headaches, eye strain, and fatigue.\\

&2. Use a monitor filter: You can install a monitor filter that blocks ultraviolet rays from your monitor, which can help reduce eye strain and fatigue.\\

&3. Talk to your doctor: If you have been experiencing symptoms of eye strain, headaches, or fatigue, talk to your doctor. He or she may recommend \\
&eye drops, glasses, or other solutions to manage your symptoms. \\
\cmidrule(r){2-2}
& Instruction: Translate the sentence ’I have no mouth but I must scream’ into Spanish.\\
& Response: No tengo boca pero debo gritar.\\
\cmidrule(r){2-2}
&Instruction: Who is Donald Trump? \\
& Response: Donald Trump is a U.S. president who was elected in 2016. He is the 45th president of the United States of America. \\
\bottomrule
\end{tabular}
}
\caption{ Generated Examples from LLaMA-v2-7b and Bolaco.}
\label{tab:generate}
\end{table*}

%% file: table/generalization.tex
\begin{table}[t]
\centering
\resizebox{0.48\textwidth}{!}{
\begin{tabular}{lcccc}
\toprule
&Wikitext ($\downarrow$) & PTB ($\downarrow$) & Zero-shot ($\uparrow$) \\
\hline
Wikipedia & 7.96 & 45.84 & 62.57 \\
Wikitext  & 7.61 & 48.37 & 62.14\\
C4 & 7.65 & 44.56 & 62.07 \\
Arxiv & 8.46 & 46.77 & 62.11 \\
\bottomrule
\end{tabular}
}
\caption{Results under different validation data.}
\label{tab:valid}
\end{table}